\RequirePackage[2020-02-02]{latexrelease} 
\documentclass{tADR2e}
\usepackage{color}
\usepackage{url}
\usepackage{comment}
\usepackage{umoline}
\usepackage[normalem]{ulem}
\usepackage{proofread} 

\begin{document}

\jvol{00} \jnum{00} \jyear{2023} \jmonth{August}


\title{Pixyz: a Python library for developing deep generative models}

\author{Masahiro Suzuki$^{a}$$^{\ast}$ \thanks{$^{\ast}$ Email: masa@weblab.t.u-tokyo.ac.jp\vspace{6pt}}, Takaaki Kaneko$^{b}$ and Yutaka Matsuo$^{a}$\\\vspace{6pt}  $^{a}${\em{The University of Tokyo, Tokyo, Japan}};
$^{b}${\em{Hiroshima University, Hiroshima, Japan}}}

\maketitle

\begin{abstract}
With the recent rapid progress in the study of deep generative models (DGMs), there is a need for a framework that can implement them in a simple and generic way. In this research, we focus on two features of DGMs: (1) deep neural networks are encapsulated by probability distributions, and (2) models are designed and learned based on an objective function. Taking these features into account, we propose a new Python library to implement DGMs called \emph{Pixyz}. This library adopts a step-by-step implementation method with three APIs, which allows us to implement various DGMs more concisely and intuitively. In addition, the library introduces memoization to reduce the cost of duplicate computations in DGMs to speed up the computation. We demonstrate experimentally that this library is faster than existing probabilistic programming languages in training DGMs.
\end{abstract}

\begin{keywords}
deep generative models, Python library
\end{keywords}\medskip

\section{Introduction}
Artificial intelligence has made great progress in recent years through deep learning~\cite{lecun2015deep}.
Among the various trends in deep learning, deep generative models (DGMs) (i.e., generative models based on deep neural networks (DNNs)) have been the focus of significant attention~\cite{kingma2013auto, rezende2014stochastic, goodfellow2014generative, dinh2016density, kingma2018glow, du2019implicit, van2016pixel, song2019generative, ho2020denoising}.
Generative models represent the process of generating data, whereas DGMs can also handle high-dimensional data with more complex dependencies that cannot be handled by conventional probabilistic models. Therefore, they can be used to learn generative models of images~\cite{radford2015unsupervised} and text~\cite{bowman2015generating}, generate them, and infer from them to obtain good representations.
In addition, they can be used for anomaly detection~\cite{schlegl2017unsupervised}, out-of-distribution detection~\cite{liu2020energy}, transformation between modalities~\cite{suzuki2016joint}, and learning larger datasets from the environment to acquire models of the environment for reinforcement learning~\cite{hafner2019learning, hafner2019dream}.

There are various types of DGMs, such as variational autoencoders (VAEs)~\cite{kingma2013auto, rezende2014stochastic}, generative adversarial networks (GANs)~\cite{goodfellow2014generative}, and flow-based models~\cite{dinh2016density, kingma2018glow}, and many models based on them have been proposed so far. Although the implementations of these studies often use deep learning libraries in the Python language, such as PyTorch~\cite{paszke2017automatic} or Tensorflow~\cite{dillon2017tensorflow}, they are implemented in their own unique ways. As a result, their code is often difficult to understand due to their unique designs. The lack of a unified implementation format is also a problem in terms of maintenance.

Probabilistic programming languages (PPLs)~\cite{van2018introduction} are known as languages for constructing ``computational graphs'' of probabilistic models and inferring the posterior distribution of arbitrary variables in those models from observed data.
They provide a unified framework for building and inferring probabilistic models. A large number of PPLs have been developed, and in recent years, attention has focused on those that run in the Python language. In particular, Pyro~\cite{bingham2018pyro}, PyMC3~\cite{salvatier2016probabilistic}, and TensorFlow Probability~\cite{dillon2017tensorflow} (including Edward~\cite{tran2016edward}) have been developed as PPLs that can build part of the computational graph of a probabilistic model in DNNs. These PPLs can benefit from Graphics Processing Unit (GPU) acceleration by using PyTorch or Tensorflow as their host language.

Despite their versatility as programming languages, PPLs are rarely used in practice as a library for implementing DGMs.
We hypothesize that this is because these existing PPLs do not take into account the characteristics of DGMs, which are different from traditional (non-deep) probabilistic models. Specifically, we focus on two inherent features of DGMs: (1) DNNs are encapsulated by probability distributions and (2) models are designed and learned based on an objective function. Our goal is to develop a library that takes these features into account and allows for a \emph{concise} and \emph{intuitive} implementation of various DGMs. Note that ``concise'' in this paper means that various models can be implemented with less code within the same framework, and ``intuitive'' means that the model to be implemented can be put into the implementation as directly as possible (from its equation) and that the correspondence between the implementation and the model is easy to understand after implementation. Note further that we do not aim to develop a novel PPL, i.e., a coherent language that encompasses all the features of the previous PPLs, but instead to develop a library dedicated to the ease of implementation (from the two perspectives mentioned above) of DGMs.

\begin{figure}[tb]
\centering
  \includegraphics[scale=0.83]{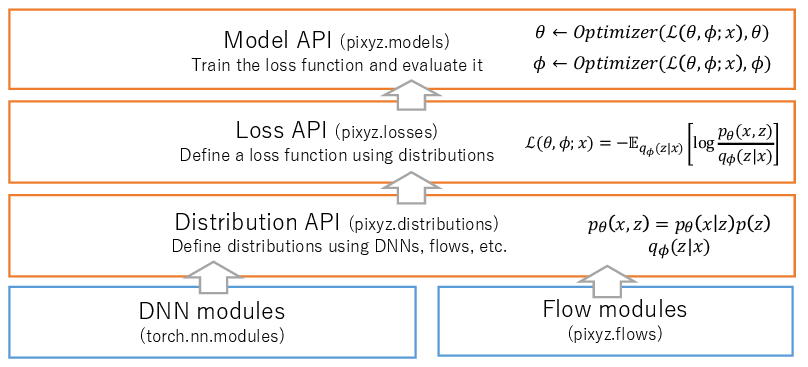}
 \caption{Implementation flow of Pixyz with three APIs.}
 \label{fig:graph}
\end{figure}

We propose a novel DGM framework called Pixyz\footnote{\url{https://pixyz.io}}, which is based on Python and PyTorch. This introduces a step-by-step implementation scheme according to three new APIs: Distribution API, Loss API, and Model API (see Figure~\ref{fig:graph}). These APIs have a hierarchical relationship, with the higher-level API encapsulating the implementation of the lower-level API. This \emph{encapsulation} provides a framework for implementation that takes into account the above two inherent features of DGMs, allowing for concise and intuitive implementation.

Distribution API supports the construction of distribution classes by means of DNNs. Using the instances obtained from each distribution class, we can perform sampling and likelihood estimation in the same framework, regardless of the form of DNN used in them.
In addition, joint distributions can be designed by the operation of multiplying conditional distributions constructed by Distribution API. This can be viewed as a define-by-run (delayed execution)\footnote{The terms \emph{define-and-run} and \emph{define-by-run} are used in the deep learning library Chainer~\cite{tokui2019chainer}.} scheme similar to many conventional PPLs without automatic differentiation in the sense that the structure of the model is determined before the data is provided, while the DNN implemented in PyTorch, which defines the distribution, can be considered a define-by-run (sequential execution) scheme.
In other words, Pixyz employs both a \emph{flexible} definition for each probability distribution depending on the data and an \emph{explicit} design for the entire probability model before the data are given. In contrast, for PPLs with automatic differentiation, such as Pyro, the distributions and DNNs are all defined in a define-by-run scheme and the DNNs are not encapsulated, thus making the implementation of complex DGMs redundant and low readability.

Loss API allows us to intuitively define the objective function corresponding to the model being implemented from the probability distribution constructed by Distribution API. In conventional PPLs without automatic differentiation, the optimization method and the objective function are considered identical; therefore, it is difficult to freely design the objective function. Pixyz, on the other hand, alleviates this problem by separating the definition of the objective function and optimization method, with the former handled by  Loss API and the latter by Model API, an upper-layer API of Loss API. In this way, each API in Pixyz is designed to consider the different characteristics of DGMs, thereby enabling concise and intuitive\delspan{flexible} implementation and learning of DGMs.

We have also devised speedups in Pixyz.
VAEs, representative models among DGMs, require multiple inferences based on the same inference model to be performed on the objective function, which increases the computational cost of the objective function if implemented in a straightforward manner.
Pixyz utilizes caching to memoize the results of inference once performed on an input and use them for the second and subsequent inference runs on the same input, thereby achieving high-speed processing.

The paper is organized as follows. We first explain the difference between DGMs and conventional probabilistic generative models, and then explain the difficulties in implementing them in existing PPLs with Python and automatic differentiation frameworks. Next, we introduce Pixyz and explain how it alleviates this difficulty. We then explain the learning acceleration method by memoization on Pixyz. In our experiments, we compare the execution time of DGMs implemented in Pixyz and Pyro and show that Pixyz is faster at training DGMs. 
Finally, we discuss the potential applications in robotics, the current limitations of Pixyz, and some future perspectives.

\section{Background: Deep Generative Models}
In the generative model framework, the given data $\mathbf{x}$ is assumed to be generated from the true data distribution $p_{data}(\mathbf{x})$, and a probabilistic model $p_{\theta}(\mathbf{x})$ (where $\theta$ is a parameter of the probabilistic model) is learned to approximate it. This probabilistic model is called a generative model because it models the generative process of the observed variables.

In general, a generative model assumes that the latent variable $\mathbf{z}$ is an unobserved factor behind the observed variable; that is,
\begin{equation}
p_{\theta}(\mathbf{x}) = \int p_{\theta}(\mathbf{x}, \mathbf{z})  d\mathbf{z}.
\end{equation}
In such a generative model with multiple random variables, conditional dependence between different variables, which we call \emph{design of generative models} in this paper, is assumed.
For example, we assume that a generative model with an observed variable $\mathbf{x}$ and a latent variable $\mathbf{z}$ is 
\begin{equation}
p_{\theta}(\mathbf{x}, \mathbf{z})=p_{\theta}(\mathbf{x}|\mathbf{z})p_{\theta}(\mathbf{z}),
\label{eq:vae}
\end{equation}
which means that we design the model with a conditional distribution $p_{\theta}(\mathbf{x}|\mathbf{z})$ and a distribution $p_{\theta}(\mathbf{z})$\footnote{The factorization of the joint distribution in Eq.~\ref{eq:vae} always follows from the chain rule, but we say ``design'' here in the sense that the joint distribution is defined only by the factorization of $p_{\theta}(\mathbf{x}|\mathbf{z})$ and $p_{\theta}(\mathbf{z})$, not $p_{\theta}(\mathbf{z}|\mathbf{x})$ and $p_{\theta}(\mathbf{x})$.}.
For another example, we assume that a generative model with two observed variables $\mathbf{x}_1$ and $\mathbf{x}_2$ and a latent variable $\mathbf{z}$ is
\begin{equation}
p_{\theta}(\mathbf{x}_1, \mathbf{x}_2, \mathbf{z})=p_{\theta}(\mathbf{x}_1|\mathbf{z})p_{\theta}(\mathbf{x}_2|\mathbf{z})p_{\theta}(\mathbf{z}),
\label{eq:vae2}
\end{equation}
which corresponds to designing a model such that $\mathbf{x}_1$ and $\mathbf{x}_2$ are conditionally independent given $\mathbf{z}$. Such conditional dependence can be represented as a graphical model.

After learning the generative models, sampling can be performed to generate new data.
From the generative model defined by Eq.~\ref{eq:vae}, we can generate samples by ancestral sampling, such as $\mathbf{z}\sim p_{\theta}(\mathbf{z})$ and $\mathbf{x}\sim p_{\theta}(\mathbf{x}|\mathbf{z})$, while the generative model by Eq.~\ref{eq:vae2} is sampled as $\mathbf{z}\sim p_{\theta}(\mathbf{z})$, $\mathbf{x}_1\sim p_{\theta}(\mathbf{x}_1|\mathbf{z})$, and $\mathbf{x}_2\sim p_{\theta}(\mathbf{x}_2|\mathbf{z})$.
In addition, estimating the posterior distribution over the latent variable $p_{\theta}(\mathbf{z}|\mathbf{x})$ (or $p_{\theta}(\mathbf{z}|\mathbf{x}_1, \mathbf{x}_2)$), i.e., inference, is also an important task in generative models\footnote{In Bayesian probabilistic modeling, inference and learning are the same processes, whereas we follow the convention in deep generative modeling by referring to estimating the posterior distribution of the latent variable as {\it inference} and parameter optimization as {\it learning}.}.

Conventionally, simple probability distributions that are easy to calculate are used to construct generative models. 
However, such models cannot capture the distribution of input data with high dimensionality and complex dependencies among dimensions, such as natural images. 
Therefore, instead of dealing directly with high-dimensional data, these generative models often use these preprocessed low-dimensional features as input; thus, they are mainly used to infer latent variables from the data rather than to generate it.

DGMs can directly capture the distribution of data with such complex dependencies by representing the probability distribution in DNNs. This makes it possible to learn and generate high-resolution natural images, which was not possible with conventional generative models~\cite{radford2015unsupervised}.

Known types of DGMs include VAEs~\cite{kingma2013auto, rezende2014stochastic}, GANs~\cite{goodfellow2014generative}, and flow-based models~\cite{dinh2016density, kingma2018glow}. Compared to conventional generative models, these models have the following common features:
\begin{enumerate}
\item \emph{DNNs are encapsulated by probability distributions.}
As mentioned above, probability distributions in DGMs are represented by DNNs; however, the manner of this representation differs depending on the model. 
For example, in VAEs, the parameter $\lambda$ of arbitrary parametric probability density function $g(\mathbf{x};{\lambda})$\footnote{In VAEs, a Gaussian distribution is often selected as $g(\mathbf{x};{\lambda})$, where the parameter $\lambda$ is the mean vector and co-variance matrix of Gaussian.} is represented as the output of a DNN $f_{\theta}$ with $\mathbf{z}$ as input, which is considered to define the conditional distribution $p_{\theta}(\mathbf{x}|\mathbf{z})$:
\begin{equation}
p_{\theta}(\mathbf{x}|\mathbf{z}) \equiv g(\mathbf{x}; {\lambda=f_{\theta}(\mathbf{z})}),
\label{eq:explicit}
\end{equation}
where $\equiv$ denotes that the probability density function on the left-hand side is defined by the parametric probability density function on the right-hand side.

In GANs, the forward propagation from $\mathbf{z}$ to $\mathbf{x}$ through a DNN $f_{\theta}$ is considered a sample from the conditional distribution $p_{\theta}(\mathbf{x}|\mathbf{z})$:
\begin{equation}
\mathbf{x} \sim p_{\theta}(\mathbf{x}|\mathbf{z}) \Longleftrightarrow \mathbf{x}=f_{\theta}(\mathbf{z}),
\label{eq:gan}
\end{equation}
where $\Longleftrightarrow$ implies that the generative process on its left-hand side is defined by mapping by a DNN on its right-hand side\footnote{Generative models defined in this way without parametric distributions are called implicit generative models~\cite{mohamed2016learning}.}. 

Furthermore, in flow-based models, $p_{\theta}(\mathbf{x})$ is defined as the change of variables by an invertible transformation (called flow) $f_{\theta}$ to the base distribution $p(\mathbf{z})$ over $\mathbf{z}$:
\begin{equation}
p_{\theta}(\mathbf{x}) \equiv p(f_{\theta}(\mathbf{x}))\left|\frac{d f_{\theta}(\mathbf{x})}{d \mathbf{x}}\right|.
\label{eq:flow}
\end{equation}

\begin{figure}[tb]
\centering
  \includegraphics[scale=0.83]{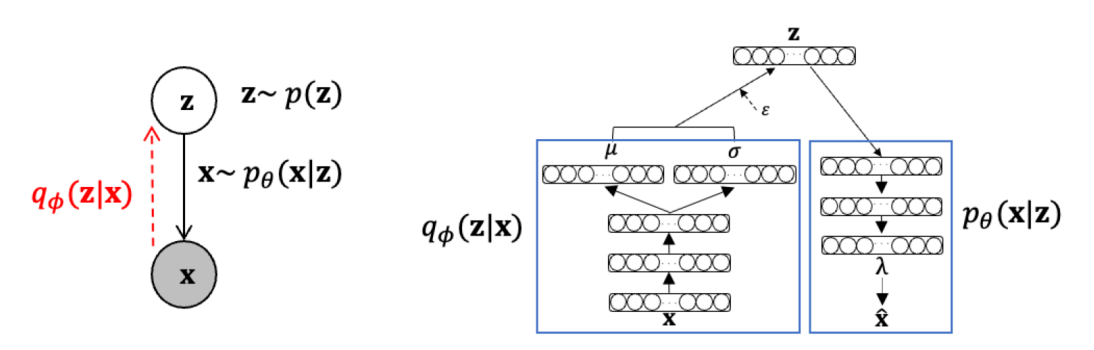}
 \caption{A graphical model of VAEs (left) and an example of DNNs that actually implement the model (right).}
 \label{fig:vae}
\end{figure}

In DGMs, probability distributions represented by DNNs are used as components to design probabilistic models. In this design process, the focus is often not on the specific structure of the DNNs, but on the relationship between the random variables.
For example, most studies of VAEs, which are latent variable models, explain the conditional dependence of observed and latent variables for each model in graphical models (Figure~\ref{fig:vae}, left). However, in actual implementations, each edge of the graphical model (i.e., each probability distribution) is represented by a neural network (Figure~\ref{fig:vae}, right), and they are often written in separate sections or appendices to the graphical model description in the paper.

\item \emph{Models are designed and learned based on the objective function.}
To learn (infer) parameters of existing complex generative models is based on sampling methods in many cases. On the other hand, all DGMs are learned by optimizing the objective function end-to-end based on backpropagation.
The objective function differs for each model; for example, the objective function of VAEs is the evidence lower bound (ELBO) of the marginal log-likelihood $\log p_{\theta}(\mathbf{x})$ of given data $\mathbf{x}$:
\begin{align}
&\mathbb{E}_{q_{\phi}(\mathbf{z}|\mathbf{x})}\left[ \log p_{\theta}(\mathbf{x}|\mathbf{z})\right] - D_{KL}[q_{\phi}(\mathbf{z}|\mathbf{x})||p(\mathbf{z})] \label{eq:elbo}\\
&= \mathbb{E}_{q_{\phi}(\mathbf{z}|\mathbf{x})}\left[ \log \frac{p_{\theta}(\mathbf{x}|\mathbf{z})p(\mathbf{z})}{q_{\phi}(\mathbf{z}|\mathbf{x})}\right]\label{eq:elbo2} \\ 
&\leq \log p_{\theta}(\mathbf{x}),\nonumber
\end{align}
where $q_{\phi}(\mathbf{z}|\mathbf{x})$ is an approximated posterior distribution (or inference model). The first term in Equation~\eqref{eq:elbo} represents the negative reconstruction error due to the inference model $q_{\phi}(\mathbf{z}|\mathbf{x})$ and the generative model $p_{\theta}(\mathbf{x}|\mathbf{z})$, and the second term represents the regularization of the inference model by the prior $p(\mathbf{z})$.
Besides that, the objective function of GANs is the Jensen--Shannon (JS) divergence between the data distribution and the generative model (which requires the introduction of a discriminative model and adversarial learning), and that of the flow-based models is the marginal log-likelihood of given data (which is equivalent to the Kullback--Leibler (KL) divergence between them). 

In addition, in DGMs, the regularization of latent variables (or representations) is often imposed as a regularization term in the objective function. For example, in VAEs, increasing the regularization term in Equation~\eqref{eq:elbo} leads to disentangled representations~\cite{higgins2016beta}.

From these facts, the definition of the objective function also plays an important role in the design of DGMs, which is another major difference from existing generative models.
\end{enumerate}

Based on these features, we next discuss the challenges in implementing DGMs in existing probabilistic modeling libraries.

\section{Related works}

\subsection{Probabilistic programming languages based on Python and automatic differentiation frameworks}
PPLs are primarily intended to automate arbitrary inference in probabilistic models, and a large number of languages have been developed. 
PPLs have been implemented based on various basic languages (e.g., PRISM is based on Prolog), BUGS, one of the early PPLs, is a self-contained language and performs inference based on Markov chain Monte Carlo (MCMC). 
Stan~\cite{carpenter2017stan}, inspired by this language, is based on C++ and supports efficient inference using Hamiltonian Monte Carlo (HMC) and its variant, the No-U-Turn Sampler.

Python has been used frequently in recent years to implement machine learning models. Therefore, several PPLs based on Python have been developed, including bayesloop and pomegranate, which allow users to design probabilistic models and perform inference in Python.
Currently, Tensorflow Probability~\cite{dillon2017tensorflow} (Edward~\cite{tran2016edward}) and Pyro~\cite{bingham2018pyro}, which are based on Python and automatic differentiation frameworks such as TensorFlow~\cite{dillon2017tensorflow} and PyTorch~\cite{paszke2017automatic}, have been actively developed.
They enable the implementation of probabilistic models with DNNs and their fast sampling, inference, and training on GPUs.

\subsection{Implementation of DGMs with PPLs}
These general-purpose PPLs with automatic differentiation frameworks can implement arbitrary deep probabilistic models and can also do so for DGMs. However, these PPLs are currently rarely used in many DGM studies. In this paper, we refer to these PPLs with automatic differentiation frameworks, including Pyro, Edward, and TensorFlow Probability, as \emph{conventional approaches} for implementing DGMs.

We believe that the implementation of DGMs using these conventional approaches does not improve code clarity and readability and that this is due to the fact that these are not designed with the aforementioned characteristics of DGMs. In particular, we focus on the two points described below.

First, to the best of our knowledge, conventional approaches treat DNNs and probability distributions equally and thus do not officially support the encapsulation of DNNs into probability distributions; thus, it is difficult for them to satisfy the first feature of DGMs.

Pyro, a PyTorch-based PPL, allows us to write the generative process of multiple random variables in a Python method, and then use methods such as \texttt{trace}, \texttt{condition}, and \texttt{block} included in \texttt{pyro.poutine} to calculate the likelihood of its joint distribution and its conditional distribution given arbitrary variables. 
In a sense, these functions realize the concealment referred to above. However, the form of the instances generated by them differs from that of the instances of the original probability distribution class, so they can no longer be sampled or their likelihood estimated in the same way. If the joint probability distributions are designed such that $p(\mathbf{x},\mathbf{y})=p(\mathbf{x}|\mathbf{y})p(\mathbf{y})$, we believe that sampling and likelihood calculations from $p(\mathbf{y})$, $p(\mathbf{x}|\mathbf{y})$, and $p(\mathbf{x},\mathbf{y})$ should be treated in the same way. 
There is also a concern that the above Pyro implementation of joint and conditional distributions may be somewhat counterintuitive in the sense that multiple additional methods are required.

Tensorflow Probability, a Tensorflow-based PPL, includes the \texttt{JointDistributionSequential} class~\cite{piponi2020joint} to solve the problem of defining joint distributions, which generates joint distribution instances with a list of generative processes as arguments. These instances can perform sampling and likelihood estimation in the same way as other probability distributions. This advantage is similar to the Pixyz distribution API proposed in this paper; however, Pixyz is more intuitive because it can construct the joint distribution as a \emph{product} of distributions encapsulating DNNs\footnote{Furthermore, we would like to emphasize that Pixyz, which introduced this API for defining joint distributions, was first released  (06/10/2019) before the publication of Tensorflow Probability's paper about \texttt{JointDistributionSequential}~\cite{piponi2020joint}.}.

Second, many PPLs do not support the flexible definition of objective functions, which means they are difficult to satisfy the second feature. This is because, as discussed in Section 2, the design and learning of DGMs differs greatly from  those of existing probabilistic models, and many probabilistic modeling libraries support the implementation of such general probabilistic models. 
Although some deep probabilistic modeling libraries~\cite{tran2016edward,bingham2018pyro} support implementations such as VAEs and GANs, they are pre-designed as classes for learning, and thus their objective functions cannot be defined flexibly.

Probabilistic Torch (ProbTorch)~\cite{siddharth2017learning} is the closest library to Pixyz proposed in this study. 
ProbTorch is a library dedicated to the implementation of DGMs, and can be implemented considering the first feature of DGMs, which are common to Pixyz.
However, Pixyz has the great advantage of being able to directly implement products of probability distributions, objective functions, etc., which ProbTorch lacks. In addition, while ProbTorch is mainly designed to implement VAEs, Pixyz can easily implement not only VAEs but also GANs, flow-based models, and complex DGMs that combine them.

\section{Proposed library: Pixyz}
In this section, we describe our proposed library, Pixyz, which is based on Python and uses PyTorch to design DNNs, takes samples from probability distributions, calculates their likelihoods, and optimizes them.

Pixyz consists of a step-by-step framework with three APIs.
As shown in Figure~\ref{fig:graph}, the implementation flow is as follows:
(1-1) define probability distributions by the neural networks ({\bf Distribution API}), 
(1-2) design a probabilistic model by calculating the product of probability distributions ({\bf Distribution API}),
(2) design the objective function ({\bf Loss API}), and
(3) learn to optimize the defined objective function ({\bf Model API}). Figure~\ref{fig:vae} shows a concrete implementation example of VAEs (Equation~\ref{eq:elbo}) that supplements the explanation of each API.

The advantage of this step-by-step implementation by APIs is that they do not interfere with each other, which makes for a concise and intuitive implementation. Below, we explain how each API contributes to a concise and intuitive implementation with sample implementations.

\begin{figure}[tb]
\centering
  \includegraphics[scale=0.78]{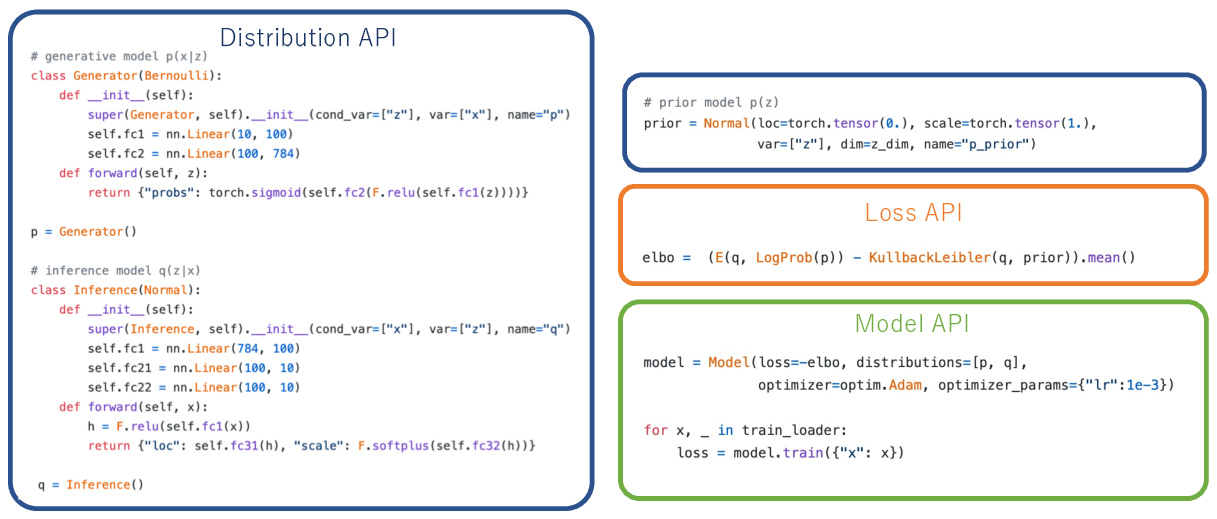}
 \caption{Example of a VAE implementation by Pixyz.}
 \label{fig:vae}
\end{figure}

\subsection{Distribution API}
The first step in Pixyz is to define the probability distributions that make up the probabilistic model. Distribution API has two major features for implementing probabilistic models: 1. it can define probability distributions in the form of encapsulating DNNs and 2. it can implement joint distributions by the product of the distributions of each variable.

\subsubsection{Defining probability distributions by the neural networks}
Basic probability distributions such as Gaussian (\texttt{Normal}) and Bernoulli (\texttt{Bernoulli}) distributions are implemented under \texttt{pixyz.distributions}.
The API for these distribution classes is almost identical to that of the PyTorch Distribution\footnote{\url{https://pytorch.org/docs/stable/distributions.html}}, and the arguments of these classes include the parameters of their corresponding distributions. 
It is also possible to set the name of the random variable in \texttt{var} and the name of the probability distribution in \texttt{name} as their arguments. 
To design a conditional distribution, it can be initialized by specifying the name of the random variable to be conditioned as the initial value of \texttt{cond\_var}.

After creating an instance of the distribution class in this manner, methods such as \texttt{sample} and \texttt{get\_log\_prob} can be used to sample and evaluate the likelihood of a given sample, respectively. In Pixyz, all samples are represented in dictionary form (\texttt{dict}), where user-defined random variable names and their realizations are stored as keys and values, respectively.

Next, we explain how to define a probability distribution with a DNN in Pixyz. 
As shown in Equation~\eqref{eq:explicit}, when defining the parameter $\lambda$ of a distribution $g$ as the output of a neural network $f_{\theta}$ with $\mathbf{z}$ as input, we first inherit the class that corresponds to the distribution $g$, and then, as in the usual PyTorch implementation, we write the network $f_{\theta}$ in its constructor and the computation flow from input $\mathbf{z}$ to the output (i.e., the parameter of the distribution $\lambda$) in its \texttt{forward} method. See Figure~\ref{fig:vae} for an example of how to define generative and inference models in VAEs using DNNs. 
Equations~\eqref{eq:gan} and~\eqref{eq:flow} can also be implemented by Distribution API, which are described in Section 5.

\subsubsection{Designing a probabilistic model by calculating the product of probability distributions of each variable}

In addition, Distribution API has a feature that allows the \delspan{intuitive} multiplication of distributions. For example, if we define the probability distributions $p(\mathbf{x}|\mathbf{z})$ and $p(\mathbf{z})$, the joint distribution is their product, $p(\mathbf{x},\mathbf{z}) = p(\mathbf{x}|\mathbf{z})p(\mathbf{z})$, which also appears in Equation~\eqref{eq:elbo2}.

Pixyz allows us to define the joint distribution as a product of instances of distribution classes\footnote{At this time, different distribution classes in the product that have the same variable name are treated as the same variable.}. 
After performing the product calculation, a graphical model defining the dependencies between the variables is constructed inside the instance of the joint distribution. This graphical model is constructed using NetworkX~\cite{hagberg2008exploring}, a library for designing and computing graph networks\footnote{
Implementation-wise, joint distribution is managed as a graph by NetworkX. Each random variable with a unique name is managed as a \emph{node}, and each distribution is managed as a \emph{directed edge} with a list of its random variables and conditional random variables (usually multiple conditional variables to a single variable).
Generating samples from joint distribution can be implemented with an ancestral sampling procedure that prepares the condition variables in order by topological sorting from the connection relations.
This process outputs a dictionary with the random variable names as keys and the samples as values, aggregated along the directed edges of the graph. The log-likelihood of the joint distribution \texttt{get\_log\_prob} is computed by adding the log-likelihoods of each edge (distribution).}.
Since this joint distribution instance is also a distribution class, sampling and likelihood estimation can be performed in exactly the same framework regardless of its distribution type.

\begin{figure}[tb]
\centering
  \includegraphics[scale=0.78]{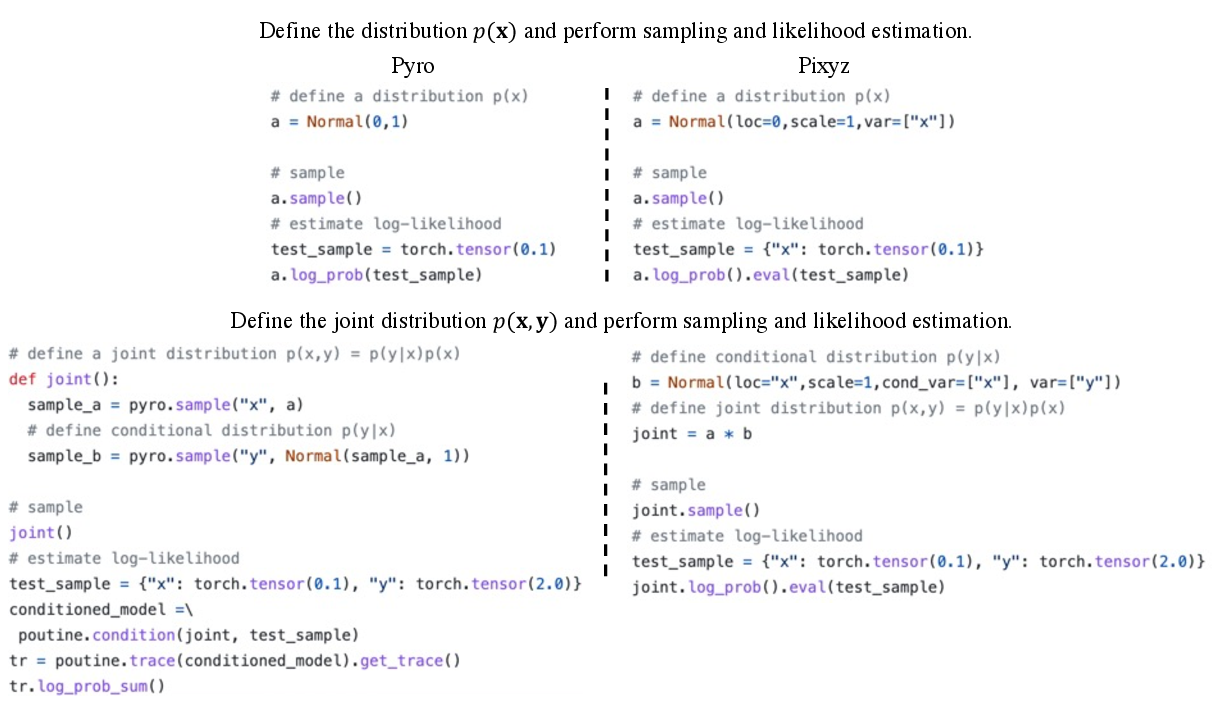}
 \caption{Comparison of Pixyz and Pyro implementations of sampling and likelihood estimation from simple and joint distributions}
 \label{fig:pixyzpyro}
\end{figure}

This idea of expressing joint distributions by the product of probability distributions is quite different from the conventional PPLs (i.e., describing generative processes in a sequential manner).
Although the approach for describing generative processes is intuitive, it only represents the flow of samples, and therefore requires a different procedure when calculating the likelihood of joint or conditional distributions.
Specifically, one must actually input the data, trace the flow of the graph, store the realizations for each random variable, calculate the likelihood of the corresponding probability distributions, and finally calculate their product.
This is the reason Pyro requires auxiliary methods such as \texttt{trace}, \texttt{condition}, and \texttt{block} for likelihood calculation, as described in Section 3. Figure~\ref{fig:pixyzpyro} shows how Pyro and Pixyz implement a joint distribution. We find that Pyro performs likelihood estimation in a special way that is different from the simple distribution, while Pixyz can be implemented in a manner consistent with a simple distribution.

On the other hand, Pixyz, which represents joint distributions by products of distribution instances, allows the structure of the graph to be defined before the data flow. 
Furthermore, since each element of the product is a distribution instance (DNNs are encapsulated within them and do not appear in the product here) and the result of the product is also a distribution instance, both the likelihood estimation and sampling from it can be easily performed without the need for additional methods. 

This difference in the way Pixyz and many MCMC-based PPLs define joint distributions can be viewed as the difference between a define-and-run (delayed execution) scheme and a define-by-run (sequential execution) scheme\footnote{The terms \emph{define-and-run} and \emph{define-by-run} are used in the deep learning library Chainer~\cite{tokui2019chainer}.}.
Since graphical models are usually \emph{designed} before data are given in probabilistic models, we believe that the new framework we have proposed is more feasible  for implementing DGMs.
Note that the samples inside each distribution are implemented by PyTorch with DNNs, so it is a define-by-run scheme. Therefore, Pixyz uses a scheme that both \emph{flexibly} designs each probabilistic distribution depending on the data and \emph{explicitly} designs the probabilistic model before the data are given.

\begin{figure}[tb]
\centering
  \includegraphics[scale=0.8]{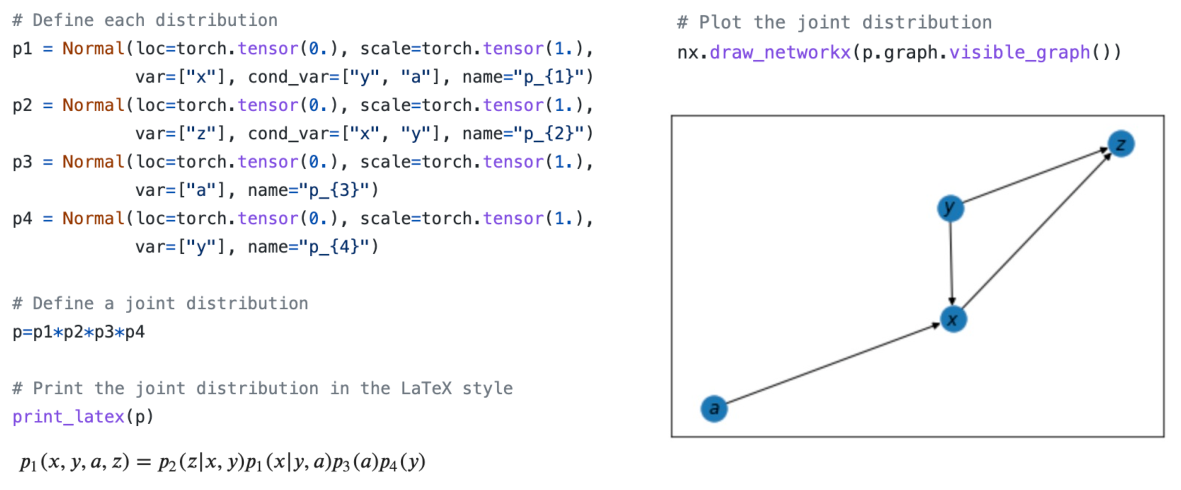}
 \caption{Definition of joint distribution by Pixyz. We can visualize it as a LaTeX-style equation (left) or draw it as a NetworkX graph (right).}
 \label{fig:joint_prob}
\end{figure}

One might be concerned that if distribution instances are abstracted in this manner, we might lose track of the distributions and networks that  designed them. In Distribution API, we can use Python's \texttt{print} method to check the distribution instance information. 
In addition, the \texttt{print\_latex} method included in Pixyz can be used to display the equation of a distribution instance in LaTeX format (Figure~\ref{fig:joint_prob}, left). If this distribution instance is joint distribution, the factorization of the distributions that compose it is displayed.
Furthermore, as mentioned above, joint distribution is treated internally as a graph of NetworkX, so it can also be visualized as a graph (Figure~\ref{fig:joint_prob}, right).

\subsection{Loss API}
Pixyz has a higher-level API called Loss API that calculates the objective function from given distributions.
Loss API includes log-likelihood, entropy, expectation, divergence between distributions, adversarial learning loss, etc. in \texttt{pixyz.losses}. 
Each of these takes distribution instances as arguments at initialization and generates an instance of the objective (or loss) function.

A feature of Loss API is that it can perform arbitrary arithmetic operations between objective function instances, whereas Distribution API can multiply distributions. 
For example, the objective function of VAEs in Equation~\eqref{eq:elbo} is composed of two terms. Therefore, we can define this objective function by creating an objective function instance for each of them and summing them. In recent complex DGMs, the setting of the objective function itself plays an important role in the design of the model~\cite{gregor2018temporal} and as a result, their objective functions are now composed of multiple terms.
As mentioned above, the implementation of such models is difficult with conventional PPLs. On the other hand, with Loss API, Pixyz makes it possible to implement arbitrary DGMs by simply writing the equation for their objective function. For the implementation of VAE using the Loss API, please refer to the box named Loss API in Figure~\ref{fig:vae}.
In addition, as with Distribution API, the \texttt{print} and \texttt{print\_latex} methods allow us to check what equation we have defined for the objective function instance.

The value of the objective function instance can be evaluated by providing the data that corresponds to the observed variable; Loss API allows the evaluation of the instance by setting the data as the argument of the \texttt{eval} method.
Thus, the framework of Loss API, in which an equation is first defined and then the values are evaluated, can be considered a define-and-run scheme (delayed execution), similar to the likelihood calculation and sampling in Distribution API.

Another advantage of Loss API is that it separates the definition of the objective function from the learning algorithm. For example, in libraries such as Edward~\cite{tran2016edward}, the model (objective function) and the learning algorithm are integrated, thus making it difficult to extend to complex DGMs. Meanwhile, Loss API in Pixyz allows us to create arbitrary objective functions; for example, we can implement models that combine VAEs and adversarial learning (see Section 5.1 for detail). 

\subsection{Model API}
Model API creates a model instance by initializing the model class (\texttt{pixyz.models.Model}) and passing it the objective function instance defined in Loss API, the distribution instance to be learned, and the optimization algorithm. By passing data to the \texttt{train} and \texttt{test} methods of this instance and executing them, we can train and test the model.
In addition to \texttt{pixyz.models.Model}, which allows users to set arbitrary objective functions, there are model classes with internally defined objective functions for users who want to train basic DGMs such as VAEs and GANs (\texttt{pixyz.models.VAE} and \texttt{pixyz.models.GAN}).

\subsection{Toward faster execution: memoizing inference results}

\begin{figure}[tb]
\centering
  \includegraphics[scale=0.75]{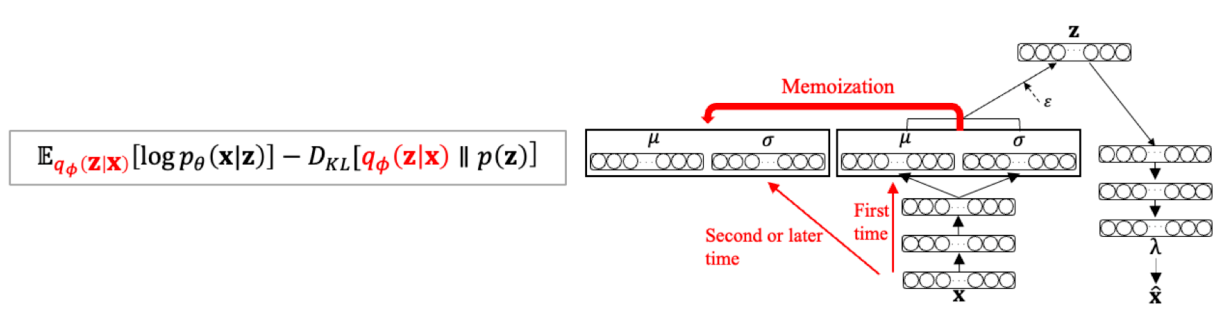}
 \caption{Inference by the same DNN given the same input appearing twice in both terms of the objective function (ELBO) of VAEs (left). 
 We propose a faster evaluation of the objective function by memoization, i.e., memoizing the input forward propagated parameters and using the memoized parameters the second or later times the same input is given (right).}
 \label{fig:memoization}
\end{figure}

When evaluating an objective function of DGMs, the same calculation might be performed multiple times in the equation. For example, in the objective function for VAEs in Equation~\eqref{eq:elbo}, an inference model given the same input appears in each term (Figure~\ref{fig:memoization}, left). Therefore, if implemented in a straightforward manner, the forward propagation of DNNs, in which the parameters of the inference model are calculated internally, must be performed twice, which increases the computational cost.

However, the values of the distribution parameters obtained from the same input should be the same as long as no probabilistic process (such as dropout) in the forward propagation of the DNN that defines this distribution\footnote{Note that the same values are stated here for the parameters of the inference model, not for the sampling results. The sampled results will naturally vary each time, even with the same inputs}. 
Therefore, we propose speeding up the evaluation of the objective function by \emph{memoization}.
Specifically, the output (parameter values) of the DNNs corresponding to a given input is memoized. If the input is the same at the next forward propagation, the memoized value is output. If it is different, the output result of another forward propagation overwrites the memoized value (Figure~\ref{fig:memoization}, right). To implement this memoization method, we used \texttt{lru\_cache}, which executes the Least Recently Used (LRU) cache provided by PyTorch and Python\footnote{To be precise, Python has the LRU cache functionality, and PyTorch uses that functionality to store PyTorch Tensor.}.
Memoization is built into Distribution API and is performed automatically (whether to perform memoization can be set by changing the size of the cache).

\section{Implementation Examples}
This section shows how various DGMs other than the simple VAE described in Section 4 can be implemented using Pixyz.

\begin{figure}[tb]
\centering
  \includegraphics[scale=0.8]{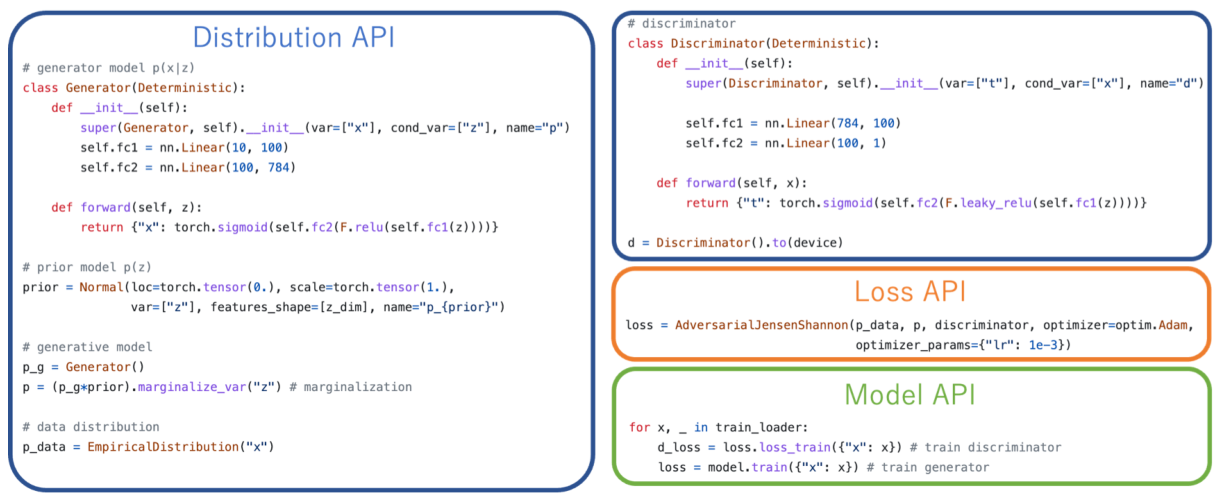}
 \caption{Example of a GAN implementation by Pixyz.}
 \label{fig:gan}
\end{figure}

\subsection{Generative adversarial networks}
Figure~\ref{fig:gan} shows the implementation of GAN using Pixyz. Firstly, the generator $p_{\theta}(\mathbf{x}|\mathbf{z})$ and discriminator $p_{\phi}(d|\mathbf{x})$ (where $d$ is an observed variable that takes on binary values) are implemented using the Distribution API. The generator is a conditional distribution represented by a DNN in the form of Equation~\eqref{eq:gan}, which can be implemented by inheriting from the \texttt{Deterministic} class. In figure~\ref{fig:gan}, both the generator and the discriminator inherit the \texttt{Deterministic} classes, i.e., they do not assume an explicit parametric probability density function.

The marginal distribution $p_{\theta}(\mathbf{x})$ is obtained by marginalizing the joint distribution $p_{\theta}(\mathbf{x}, \mathbf{z})$, which is the product of the conditional distribution and the prior distribution, with respect to the latent variable, i.e., $p_{\theta}(\mathbf{x})=\int p_{\theta}(\mathbf{x}, \mathbf{z}) d\mathbf{z} = \int p_{\theta}(\mathbf{x}|\mathbf{z}) p(\mathbf{z}) d\mathbf{z}$.
In Pixyz, the method \texttt{marginalize\_var} of the Distribution API can be used to represent the marginalized distribution for any variable as a new Distribution class. \footnote{Sampling from the marginalized distribution ( \texttt{marginalize\_var}) is implemented by selecting and returning only the samples corresponding to the probability variable of the marginal distribution from the results of ancestral sampling from the original joint distribution. For example, a sample from $p_{\theta}(\mathbf{x})=\int p_{\theta}(\mathbf{x}, \mathbf{z}) d\mathbf{z}$ is obtained by first obtaining $\mathbf{x}$ and $\mathbf{z}$ samples by $p_{\theta}(\mathbf{x}, \mathbf{z})$, and then selecting only $\mathbf{x}$ as the return value. However, likelihood computation for the marginal distribution is not currently supported.}

Then, using the Loss API, we implement the JS divergence between the true distribution $p_{true}(\mathbf{x})$ and the generative model $p_{\theta}(\mathbf{x})$. The true distribution is implemented using the \texttt{EmpiricalDistribution} class, which is a kind of dummy distribution that outputs the given input as a sample. The JS divergence can be implemented using the \texttt{AdversarialJensenShannon} class, which calculates an upper bound on the JS divergence using an optimized discriminator $p_{\phi^*}(d|\mathbf{x})$, which is the objective function of the GAN:
\begin{align}
&2 \cdot D_{JS}[p_{\theta}(\mathbf{x})||p_{true}(\mathbf{x})] + 2 \log 2 = \mathbb{E}_{d\sim p_{\phi^*}(d|\mathbf{x}),\mathbf{x}\sim p_{\theta}(\mathbf{x})}[\log d] + \mathbb{E}_{d\sim p_{\phi^*}(d|\mathbf{x}),\mathbf{x}\sim p_{true}(\mathbf{x})}[\log (1-d)] \nonumber\\
&\geq D_{JS}[p_{\theta}(\mathbf{x})||p_{true}(\mathbf{x})],
\end{align}
where
\begin{align}
\phi^* = \arg\max_{\phi} \mathbb{E}_{d\sim p_{\phi}(d|\mathbf{x}),\mathbf{x}\sim p_{\theta}(\mathbf{x})}[\log d] + \mathbb{E}_{d\sim p_{\phi}(d|\mathbf{x}),\mathbf{x}\sim p_{true}(\mathbf{x})}[\log (1-d)].
\end{align}
In addition to the JS divergence, the KL divergence (\texttt{AdversarialKullbackLeibler}) and the Wasserstein distance (\texttt{AdversarialWassersteinDistance}) have been implemented for this adversarial loss\footnote{The equations that these implement using the discriminator are listed in the appendix.}. Note that in adversarial loss, the likelihood of each distribution need not be computable, i.e., it is implicit. What makes these Loss classes unique is that they have a method for learning a discriminator called \texttt{loss\_train}. Such an implementation form makes it intuitive that learning a discriminator is for estimating the correct divergence between implicit distributions (see Model API in Figure~\ref{fig:gan}).

The advantage of this implementation method is that the objective function of adversarial learning can be implemented as \emph{the distance (divergence) between distributions}. This is not supported by other PPLs such as Pyro\footnote{Edward 1~\cite{tran2016edward} allows adversarial learning based on implicit divergence using such as \texttt{ImplicitKLqp}. However, this class cannot be described with other loss functions like Pixyz's Loss API because the objective and the optimization are integrated.}. Thus, implicit distances between arbitrary distributions can be implemented exactly as in the case of GANs. For example, FactorVAE~\cite{kim2018disentangling} minimizes an implicit KL divergence between the inference model and a shuffled inference model for each dimension in the latent space of a VAE, in addition to the VAE loss, to encourage disentanglement. Even in this case, with \texttt{AdversarialKullbackLeibler}, we can easily combine it with the VAE objective. The figure shows the Loss API part of Pixyz's implementation of FactorVAE. As can be seen, the objective function of FactorVAE, which combines reconstruction and adversarial learning, can be implemented intuitively and concisely.

\begin{figure}[tb]
\centering
  \includegraphics[scale=0.8]{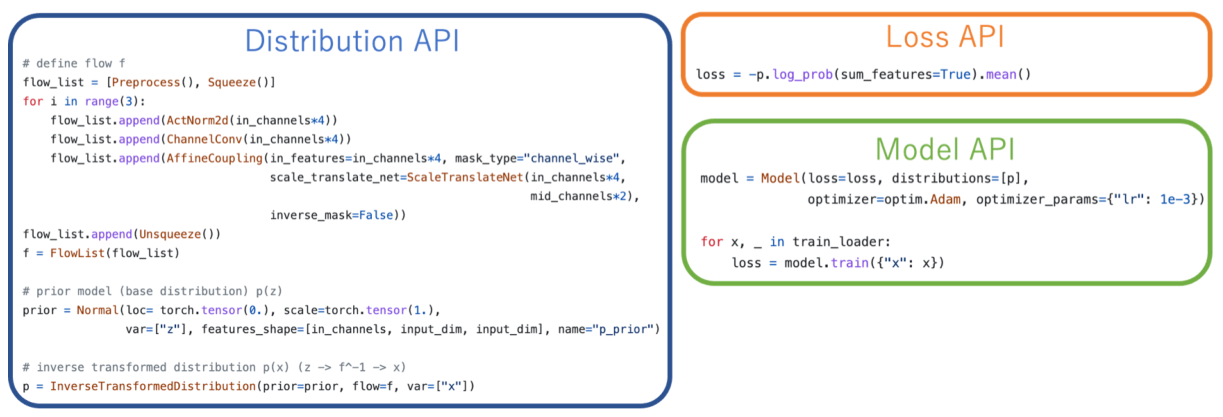}
 \caption{Example of a Glow implementation by Pixyz.}
 \label{fig:glow}
\end{figure}

\subsection{Flow-based models}
Figure~\ref{fig:glow} shows an example of Glow~\cite{kingma2018glow}, a flow-based model that implements complex transformations between variables by synthesizing reversible transformations. In Pixyz, reversible transformations are implemented in \texttt{pixyz.flows}, and a series of flows are implemented by adding these transformations to a list. In the implementation shown in this figure, the prior distribution of the latent variable $p(\mathbf{z})$ is set as the base distribution, and a flow network $\mathbf{z}=f_{\theta} (\mathbf{x})$ is defined. 

To implement the marginal distribution $p_{\theta} (\mathbf{x})$ using the above as in Equation~\eqref{eq:flow}, the \texttt{InverseTransformedDistribution} class of the Distribution API can be used. The "\texttt{Inverse}" in the class name is because the flow defined is from $\mathbf{x}$ to $\mathbf{z}$, while the process of generating the marginal distribution $p_{\theta} (\mathbf{x})$ is from $\mathbf{z}$ to $\mathbf{x}$. On the other hand, if $p(\mathbf{x})$ is a prior distribution and the flow $\mathbf{z}=f_{\theta} (\mathbf{x})$ is used to implement the joint distribution $p_{\theta} (\mathbf{z})$, the \texttt{TransformedDistribution} class is used\footnote{We typically use \texttt{InverseTransformedDistribution} for high-resolution image generative models~\cite{kingma2018glow} and \texttt{TransformedDistribution} for more complex inferences such as normalizing flow in variational inference~\cite{rezende2015variational}.}. Although the way they are defined differs greatly from the Gaussian and other distribution classes we have seen so far, the instances generated from each distribution class can also use \texttt{sample} and \texttt{get\_log\_prob} in the same manner. In addition, both \texttt{TransformedDistribution} and  \texttt{InverseTransformedDistribution} have \texttt{forward} and \texttt{inverse} methods to perform transformations in both directions. 

To perform maximum likelihood estimation of the defined $p_{\theta} (\mathbf{x})$, a negative log-likelihood $-\log p_{\theta} (\mathbf{x})$ is computed in the Loss API and learned to minimize it in the Model API.

Implementing such distributions based on the base distribution and flow network can be done in a similar framework in Pyro. The significant difference, however, is that \texttt{TransformedDistribution} and \texttt{InverseTransformedDistribution} can be treated the same way as other distribution classes, allowing them to be combined with other DGMs without concern for network structure.

\begin{figure}[tb]
\centering
  \includegraphics[scale=0.8]{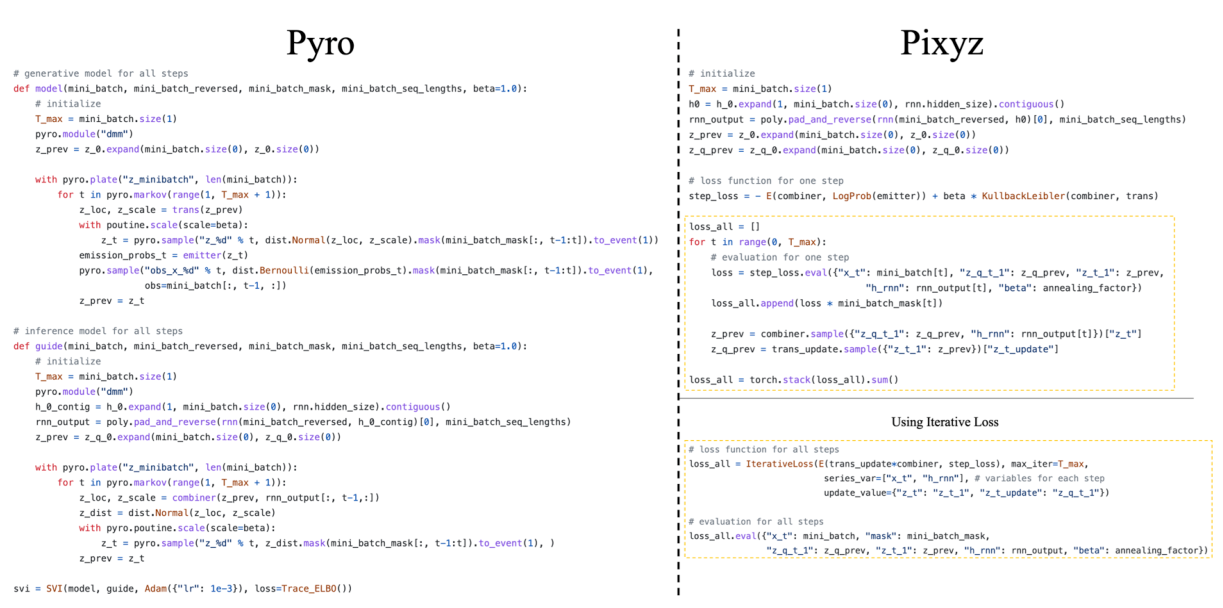}
 \caption{Comparison of Pixyz (left) and Pyro (right) implementations of the Deep Markov Model. The Pyro implementation is taken from an example in its repository, while the Pixyz implementation uses parameters that are as close as possible to the Pyro implementation. The definition of the distribution in both Pyro and Pixyz is not included, and only the definition of the loss function (in the case of Pyro, up to the point where inference and generation are defined and put into the SVI class) is provided. The lower half on the right is part of the implementation when IterativeLoss is used; when using IterativeLoss, the implementation enclosed in the yellow square in the upper half is replaced with the concise implementation in the lower half.}
 \label{fig:dmm}
\end{figure}

\subsection{Deep Markov Models (time-series DGMs)}
Finally, we present an example implementation of time series DGMs. Here we present an example implementation of the deep Markov model (DMM)~\cite{krishnan2017structured}, which can be considered a temporal expansion of VAEs. This model defines an inference model $q\left(\mathbf{z}_t \mid \mathbf{z}_{t-1}, \mathbf{x}_{t: T}\right)$, a transition model $p\left(\mathbf{z}_{t+1} \mid \mathbf{z}_t\right)$, and a generative model $p\left(\mathbf{x}_t \mid \mathbf{z}_t\right)$ for each step (where $T$ is the maximum time step). The DMM optimizes the following objective function (ELBO):
\begin{align}
\mathbb{E}_{q\left(\mathbf{z}_{1: T}\right)}\left[\log p\left(\mathbf{x}_{1: T} \mid \mathbf{z}_{1: T}\right)\right]-D_{KL}\left[q\left(\mathbf{z}_{1: T}\right) || p\left(\mathbf{z}_{1: T}\right)\right].
\end{align}

The results of the Pyro and Pixyz implementations of this model are shown in Figure \ref{fig:dmm}. In both cases, the specific implementation and optimization of each distribution are not shown, only the definition of the model. This shows that Pixyz has a more concise implementation. The most important reason for this is the difference in how the objectives are computed. In Pyro, all random variables in all steps must first be sampled, thereby setting the objective. This makes the code redundant due to the for loop, as Pyro must implement sampling every step for each distribution, and it also requires renaming the variables at each step to distinguish them. Pixyz, on the other hand, allows us first to evaluate the objective of each step and then add them together at all steps. Therefore, only one for loop is needed, and it can be written as an addition of the objects in each step, which is easy to understand. The code can be further simplified using \texttt{IterativeLoss}, which can add the loss classes in the time direction (see lower right of 
 Figure~\ref{fig:dmm}). Thus, Pixyz has the advantage of concisely implementing the time series DGMs.

For other complex DGMs implementations (Loss API), please refer to the Appendix.

\section{Experiments}
Here we compare the runtime of Pixyz with those of PyTorch and another PPL, Pyro. 
We used VAE and DMM as the models to be trained.
The architecture of each model conforms to the Pyro sample code\footnote{https://github.com/pyro-ppl/pyro/tree/dev/examples}, and the PyTorch and Pixyz implementations are aligned to the same architecture, with a batch size of 128. In this experiment, we used a computer with an Intel\textcircled{R} Xeon\textcircled{R} Gold 5218R CPU @ 2.10GHz $\times 2$ and  NVIDIA TITAN RTX as the experimental environment. 
We used the MNIST dataset~\cite{lecun2010mnist} for the VAE and the polyphonic music dataset~\cite{boulanger2012modeling} for the DMM.

Table~\ref{tab:speed} shows the VAE learning execution time for one step update. As in the experiments in \cite{bingham2018pyro}, the results are shown as averages over 10 epochs, varying the hidden layer dimension $\#h$ and the latent variable dimension $\#z$\footnote{The settings for $\#h$ and $\#z$ in the VAE experiment are the same as those in \cite{bingham2018pyro}; however, because of the different execution environment, the execution time for PyTorch and Pyro are different from those in \cite{bingham2018pyro}.}.

\begin{table}[tb]
\caption{Comparison of the execution time per step of VAE for each library (ms). For Pixyz, the VAE objective function is calculated both analytically (Analyt.; Equation~\eqref{eq:elbo}) and by Monte Carlo approximation (MC; Equation~\eqref{eq:elbo2}) and both with and without LRU cache.}
\footnotesize
\centering
\begin{tabular}{cc||cc|cccc}
\multicolumn{1}{l}{} & \multicolumn{1}{l||}{} & \multicolumn{1}{l}{} & \multicolumn{1}{l|}{} & \multicolumn{4}{c}{Pixyz}                               \\
\#z                  & \#h                   & PyTorch   & Pyro       & Analyt. w/o cache & Analyt. w/ cache & MC w/o cache & MC w/ cache \\ \hline
10                   & 400                   & 6.43±0.55 & 12.46±1.62 & 11.94±1.50   & 10.18±1.15  & 10.83±1.12   & 9.53±0.84   \\
30                   & 400                   & 6.58±0.32 & 12.02±2.04 & 11.67±1.73   & 9.63±1.21   & 10.72±1.23   & 9.56±0.84   \\
10                   & 2000                  & 6.62±0.31 & 12.47±1.59 & 11.74±1.75   & 10.39±0.95  & 10.52±1.06   & 9.67±0.65   \\
30                   & 2000                  & 6.63±0.37 & 12.58±1.67 & 12.13±1.52   & 10.25±0.96  & 10.54±1.13   & 9.27±1.04  
	\label{tab:speed}
  \end{tabular}
\end{table}

The results show that Pyro and Pixyz are both slower than PyTorch but that  Pixyz is faster than Pyro. Note that the PyTorch implementation computes the regularization term of the VAE analytically (corresponding to Equation~\eqref{eq:elbo}) while Pyro is pre-designed to compute it using a Monte Carlo approximation (Equation~\eqref{eq:elbo2}). On the other hand, Pixyz can easily implement either objective function thanks to Loss API, so we have included results for both cases.
Although the result of the analytical in Pixyz is slower than the Monte Carlo approximation, it is still faster than Pyro.
This might be due to the simpler internal operations and manipulations in Pixyz than in Pyro\footnote{Note that Pyro supports Bayesian inference of parameter distributions, sampling, etc., while Pixyz does not; therefore, one should not infer the superiority of one library over the other from these speed results alone.}.
Furthermore, Bingham et al.~\cite{bingham2018pyro} argued that the gap between Pyro and PyTorch reduces as the number of hidden layers increases, which is also true for Pixyz.
Furthermore, the speed is further improved by memoization with LRU cache. Finally, the analytic calculation and memoization showed a $20-26\%$ improvement compared to Pyro.

\begin{table}[tb]
\caption{Comparison of the execution time per step of DMM for each library (ms).}
\footnotesize
\centering
\begin{tabular}{c||c|cccc}
\multicolumn{1}{l||}{} & \multicolumn{1}{l|}{} & \multicolumn{4}{c}{Pixyz}                               \\
\#z                  & Pyro                  & Analyt. w/o cache & Analyt. w/ cache & MC w/o cache & MC w/ cache \\ \hline
20  & 583.26±78.49  & 566.50±76.32 & 529.71±77.48 & 518.36±67.32 & 473.11±71.33 \\
50  & 578.07±83.11  & 594.55±78.22 & 526.34±71.68 & 475.97±61.35 & 473.32±61.15 \\
100 & 663.44±100.48 & 552.24±65.07 & 521.08±65.42 & 521.24±64.90 & 482.46±61.01 \\
500 & 594.02±89.51  & 629.75±78.82 & 535.15±72.83 & 562.17+74.40 & 492.80±66.70
	\label{tab:speed2}
  \end{tabular}
\end{table}

Next, we compared the runtimes for a DMM.
Table~\ref{tab:speed2} shows that the trend of the difference between Pixyz and Pyro in DMM is almost the same as that in VAE. This means that the learning speed advantage of Pixyz is also valid in the time-series models.

This speed advantage of Pixyz compared to Pyro seems to be related to the manipulation of distributions, as explained above. Still, the fact that the range of improvement is not greater when comparing the results of VAE and DMM shows that the effect of improvement does not necessarily increase as the number of distributions increases. Conversely, this means that the number of distributions does not necessarily increase the effect of improvement. Conversely, since this does not affect the number of distributions so much, similar speed improvements can be expected for DGMs other than those in this experiment. However, the speedup by memoization is not expected to have the same effect for arbitrary DGMs unless the same computational path is taken multiple times when evaluating the objective function, as in the case of VAE.

\section{Discussion}
\subsection{Application to the robotics}
Robot learning aims at acquiring environmental models and decisions based on interactions with the environment. 
Because transitions in the environment involve uncertainty\footnote{To be precise, uncertainty includes two sources, aleatoric and epistemic uncertainty, and it is necessary to distinguish between them in robot learning~\cite{kroemer2021review}; however, it is important to use probabilistic models to capture it from both perspectives.}, time series probabilistic models are often used to model them~\cite{kroemer2021review}.
More recently, time-series DGMs have been used to learn representations of states (i.e., state representation learning~\cite{lesort2018state}) and transitions of them from observations obtained from the environment, which are called world models~\cite{ha2018world}. For example, Dreamer~\cite{hafner2019dream} uses recurrent state-space models~\cite{hafner2019learning}, which combine deep state-space models and deterministic RNNs, for model-based reinforcement learning.
These world models have been trained mainly on simulations~\cite{hafner2019learning,hafner2019dream,yarats2021mastering}, but have recently been applied to real-world robots to enable online learning with high sample efficiency and performance~\cite{wu2022daydreamer}. Thus, DGMs have played an important role in robot learning in recent years.

In these world models, however, we need to implement models of state transition models conditioned on actions, reconstruction models of observations via latent states, and prediction models of reward and value, all with probability distributions represented by deep neural networks, and then use these to construct objective functions. 
This can make the code complex and non-uniform when implemented naively, making it practically difficult for developers to apply those models to a variety of environments and robots.

We have shown through the display of example implementations and experimental results that Pixyz can be easily implemented in time series models, and moreover, it can be performed efficiently.
In addition to the DMMs shown above, we have confirmed that other complex time-series models such as Dreamer~\cite{hafner2019dream}, variational recurrent neural networks~\cite{chung2015recurrent}, and TD-VAE~\cite{gregor2018temporal} can be implemented with Pixyz\footnote{See the appendix for TD-VAE and the Pixyzoo repository (\url{https://github.com/masa-su/pixyzoo}) for the others.}.
Therefore, Pixyz is expected to facilitate the use of large-scale DGMs for robot learning. In recent years, it has been proposed to combine DGMs with conventional non-deep probabilistic generative models to create integrated cognitive systems, which is called Neuro-SERKET~\cite{taniguchi2020neuro}. By incorporating Pixyz into this framework, we might be able to build more complex models for robotics.

\subsection{Limitations}
We have shown that Pixyz is capable of implementing a range of deep generative models in a concise and intuitive manner. However, there are limitations to its abilities. Pixyz currently only supports ancestral sampling for generating samples and amortized variational inference for the inference method, which optimizes the ELBO with an explicitly defined inference model for latent variables. In contrast, Pyro and other PPLs support various sampling methods, such as Markov chain Monte Carlo, and inference methods for random variables and parameters using these methods. As a result, Pyro is better suited than Pixyz for implementing Bayesian deep neural networks that require parameter inference. Also, for these reasons, Pixyz is less suitable for implementing traditional non-deep generative models that use sampling methods to infer parameters, yet these can also be implemented in Pixyz using amortized variational inference. As an exception, Pixyz implements the ability to perform latent variable inference analytically in the Distribution class \texttt{MixtureModel}, which supports Gaussian distributions and can be used for EM algorithms.

Another issue is that the current Pixyz does not support models that generate samples iteratively, such as energy-based models~\cite{du2019implicit}, score-based models~\cite{song2019generative}, and diffusion models~\cite{ho2020denoising}. This is because, as mentioned above, Pixyz currently only supports generation by ancestral sampling. In the future, Pixyz will support the implementation of such models, and will be able to handle different deep generative models simultaneously.

From the perspective of representation learning, Pixyz supports representation learning as an inference of latent variables in deep generative models. For example, it can learn to acquire desirable representations by introducing constraints on the latent variables into the model's objective function, such as in beta-VAE~\cite{higgins2016beta}. However, representation learning by self-supervised learning, such as contrastive learning~\cite{chen2020simple, chen2021exploring, caron2021emerging} other than DGMs, is not supported. Nevertheless, some research has proposed reinterpreting contrastive learning as inference in DGMs~\cite{nakamura2022self}, and following this framework, Pixyz might be able to implement self-supervised learning methods.

\section{Summary and future work}
In this paper, we proposed Pixyz as a library for implementing DGMs. Pixyz utilizes three APIs to deal with the features of recent DGMs: (1) DNNs are encapsulated by probability distributions, and (2) models are designed and learned based on the objective function. We have shown that these APIs can be used to implement DGMs concisely and intuitively, and that they are also suitable for designing DGMs in terms of execution time.
In addition to the results presented in this paper, we have confirmed that DGMs learning complex environments, e.g., generative query networks~\cite{rezende2018generalized}, semi-supervised learning models using DGMs, e.g., M2 model~\cite{kingma2014semi}, other complex sequential DGMs can also be implemented and trained on Pixyz.

However, as mentioned above, there are limitations in the sense that there are many models and sampling methods that are not supported. In the future, we plan to address those aspects that Pixyz does not currently support.

DGMs and world models are more difficult to use than ordinary DNNs because of the difficulties in understanding and implementing them; we hope that  this library can play a role in eliminating such barriers.

\section*{Acknowledgements}
We would like to thank Keno Harada, Yotaro Nada, Yechan Park, Tatsuya Matsushima, and Shohei Taniguchi for their help in developing this library. 
This paper is based on results obtained from a project, JPNP16007, subsidized by the New Energy and Industrial Technology Development Organization (NEDO).
\bibliographystyle{plain}
\bibliography{tADRguide.bib}

\appendix
\section{Other complex DGMs implementations}
\begin{figure}[t]
\centering
  \includegraphics[scale=0.78]{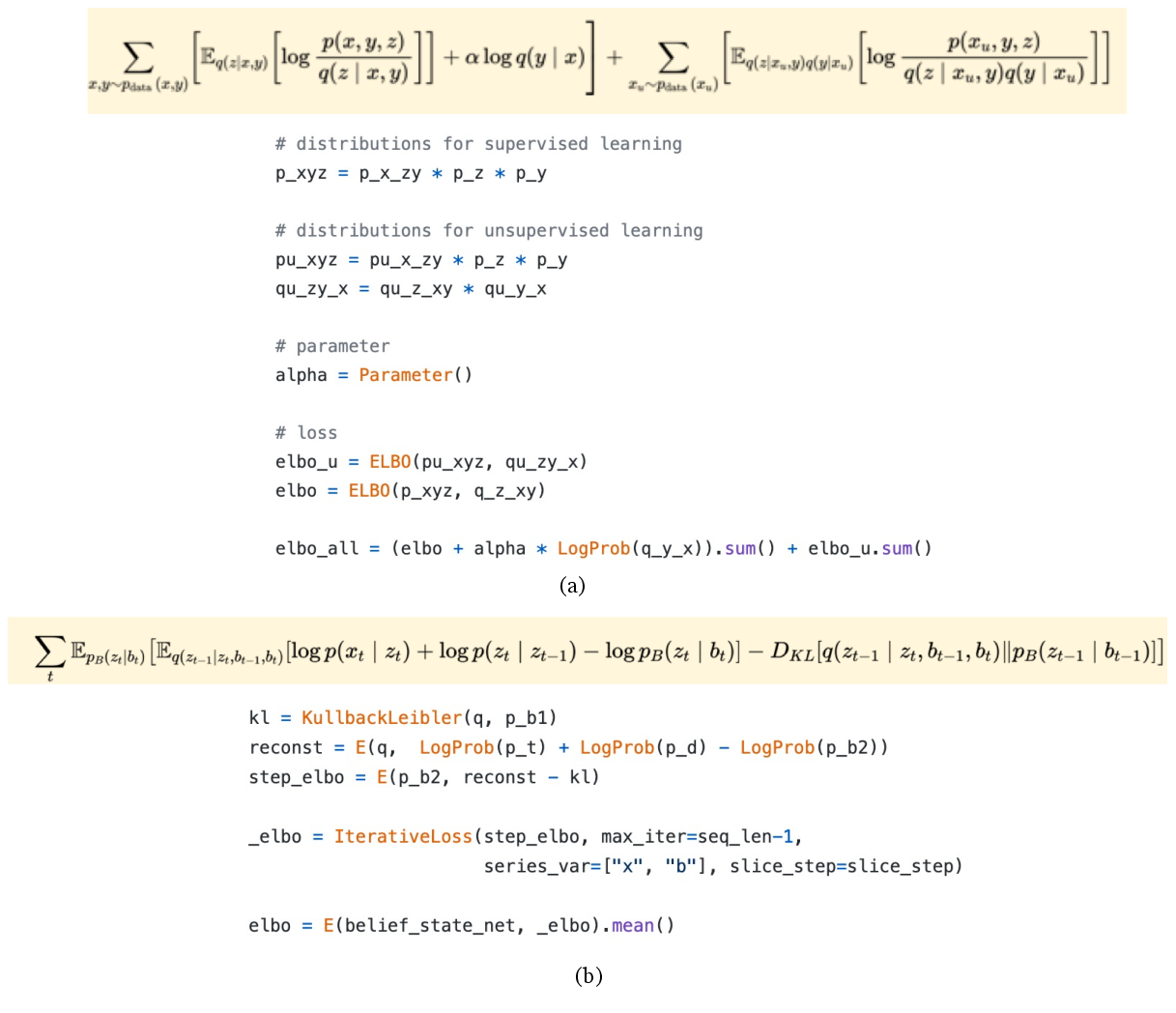}
 \caption{Example implementation of complex DGMs with the Loss API; (a): M2 model~\cite{kingma2014semi} (including the multiplication part by Distribution API), (b):TD-VAE~\cite{gregor2018temporal}. }
 \label{fig:coplex}
\end{figure}

Here we present implementations of complex DGMs using the Loss API.
Figure~\ref{fig:coplex}(a) shows the equation for the M2 model~\cite{kingma2014semi}, a semi-supervised learning model with VAEs, and its implementation using the Loss API.
It can be seen that even for models with multiple terms, the objective function can be designed using Loss API as if we wrote the equation directly. 
As a more complex model, Figure~\ref{fig:coplex}(b) shows the equation and implementation of TD-VAE~\cite{gregor2018temporal}.
Although TD-VAE is one of the complex time-series DGMs, we can confirm that it can be implemented in a concise and intuitive manner.

\section{Objective functions implemented in the adversarial loss classes}
\texttt{AdversarialKullbackLeibler} implements the following equation:
\begin{align}
&\mathbb{E}_{d\sim p_{\phi^*}(d|\mathbf{x}),\mathbf{x}\sim p(\mathbf{x})}\left[\log \frac{d}{1-d}\right] \approx\mathbb{E}_{p(\mathbf{x})}\left[\log \frac{p(\mathbf{x})}{q(\mathbf{x})}\right]=D_{KL}\left[p(\mathbf{x})||q(\mathbf{x})\right],
\end{align}
where
\begin{align}
\phi^* = \arg\max_{\phi} \mathbb{E}_{d\sim p_{\phi}(d|\mathbf{x}),\mathbf{x}\sim q(\mathbf{x})}[\log d] + \mathbb{E}_{d\sim p_{\phi}(d|\mathbf{x}),\mathbf{x}\sim p(\mathbf{x})}[\log (1-d)].
\end{align}

\texttt{AdversarialWassersteinDistance} implements the following:
\begin{align}
\sup_{||f||_{L} \leq 1} \mathbb{E}_{p(\mathbf{x})}[f(\mathbf{x})] - \mathbb{E}_{q(\mathbf{x})}[f(\mathbf{x})] =  W(p, q),
\end{align}
where
\begin{align}
d \sim p_{\phi}(d|\mathbf{x}) \Longleftrightarrow d=f_{\phi}(\mathbf{x}).
\end{align}

\end{document}